\documentclass[letterpaper, 10 pt, conference]{ieeeconf}  % Comment this line out if you need a4paper
\IEEEoverridecommandlockouts                              % This command is only needed if
\usepackage{graphics}    % for pdf, bitmapped graphics files
\usepackage{times}       % assumes new font selection scheme installed
\usepackage{amsmath}     % assumes amsmath package installed
\usepackage{amssymb}     % assumes amsmath package installed
\usepackage{graphicx}
\usepackage{algorithm}
\usepackage[noend]{algpseudocode}
\usepackage{multirow}
\usepackage{placeins}
\usepackage{siunitx}

\usepackage[font=small]{caption}

\def\figref#1{Fig.~\ref{#1}}
\def\tabref#1{Tab.~\ref{#1}}
\def\eqref#1{Eq.~(\ref{#1})}

\newcommand{\image}[1]{\mathcal{I}_\text{#1}}

\usepackage{fancyhdr}
\title{\LARGE \bf Real-time Semantic Segmentation of Crop and Weed for Precision
Agriculture Robots Leveraging Background Knowledge in CNNs}

\author{Andres Milioto \and Philipp Lottes \and Cyrill Stachniss% <-this % stops a space
  \thanks{All authors are with the University of Bonn, Germany. This work has partly 
  been supported by the EC under the grant number H2020-ICT-644227-Flourish. 
  }%
}

\begin{document}
\maketitle
\thispagestyle{fancy}
\pagestyle{fancy}

%%%%%%%%%%%%%%%%%%%%%%%%%%%%%%%%%%%%%%%%%%%%%%%%%%%%%%%%%%%%%%%%%%%%%%%%%%%%%%%%
\begin{abstract}
  Precision farming robots, which target to reduce the amount of herbicides
  that need to be brought out in the fields, must have the  ability to
  identify crops and weeds in real time to trigger weeding actions. In this
  paper, we address the problem of CNN-based semantic segmentation of crop
  fields separating sugar beet plants, weeds, and background solely based on
  RGB data. We propose a CNN that exploits existing vegetation indexes and
  provides a classification in real time. Furthermore, it can be effectively
  re-trained to so far unseen fields with a comparably small amount of
  training data. We implemented and thoroughly evaluated our system on a real
  agricultural robot operating in different fields in Germany and Switzerland.
  The results show that our system generalizes well, can operate at around
  20\,Hz, and is suitable for online operation in the fields.
\end{abstract}

%%%%%%%%%%%%%%%%%%%%%%%%%%%%%%%%%%%%%%%%%%%%%%%%%%%%%%%%%%%%%%%%%%%%%%%%%%%%%%%%
\section{Introduction}
\label{sec:intro}

% %% WHY 
Herbicides and other agrochemicals are frequently used in crop production but
can have several side-effects on our environment. Thus, one big challenge for
sustainable approaches to agriculture is to reduce the amount of agrochemicals
that needs to be brought to the fields.
Conventional weed control systems treat the whole field  uniformly with the
same dose of herbicide.  Novel, perception-controlled weeding systems offer
the potential to perform a treatment on a per-plant level, for example by
selective spraying or mechanical weed control. This, however, requires a plant
classification system that can analyze image data recorded in the field in
real time and labels individual plants as crop or weed.

\begin{figure}[]
\vspace{2mm}
\setlength\tabcolsep{0.5pt}
  \begin{tabular}{ll}
      {\includegraphics[width=0.60\linewidth]{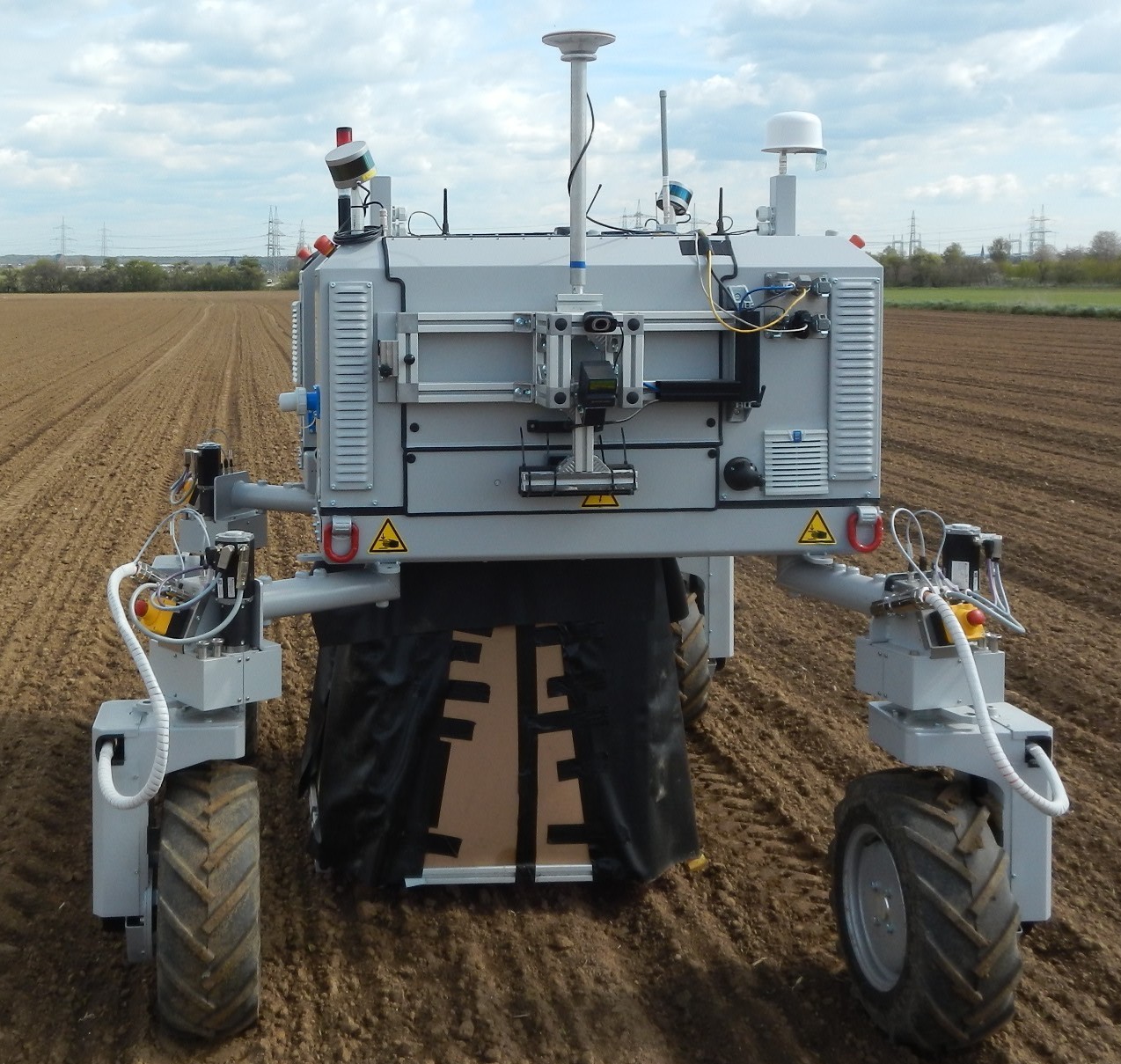}}
   & \raisebox{2.3cm}{
   \begin{tabular}{ll}
        {\includegraphics[width=0.37\linewidth]{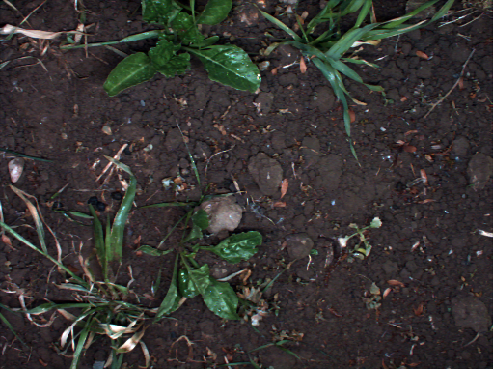}}
     \\
        {\includegraphics[width=0.37\linewidth]{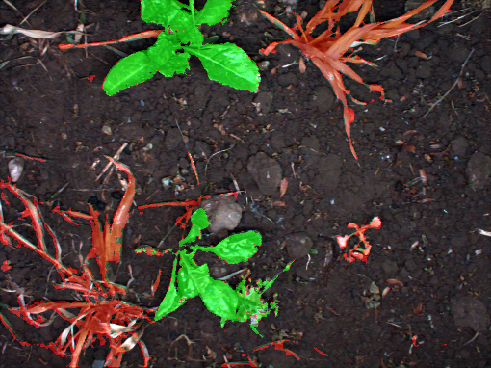}}
    \end{tabular}}
  \end{tabular}
\caption{Left: Bonirob system, used for data acquisition and field deployment. 
Right: RGB image (top) captured during field test and our corresponding classification result with crops colored green and weeds colored red (bottom).}
\label{fig:motivation}
\end{figure}

% %% WHICH PROBLEM 
In this paper, we address the problem of classifying standard RGB images
recorded in the crop fields and identify the weed plants, roughly at the
framerate of the camera. This information can in turn be used to perform
automatic and targeted weed control or to monitor fields and provide a status
report to the farmer without human interaction. There exists several
approaches to vision-based crop-weed classification. A large number of
approaches, especially those that yield a high classification performance, do
not  only rely on RGB data but furthermore require additional spectral cues
such as near infra-red information.  Often, these techniques also rely on a
pre-segmentation of the vegetation as well as on  a large set of hand-crafted
features. In this paper, we target RGB-only crop-weed classification without
pre-segmentation and  are especially interested in computational efficiency
for real-time operation on a mobile  robot, see \figref{fig:motivation} for an
illustration.

% %% MAIN CONTRIBUTION & WHAT FOLLOWS FROM THAT
The main contribution of this work is a new approach to crop-weed
classification using RGB data that relies on convolutional neural networks
(CNNs). We aim at feeding additional, task-relevant background knowledge to
the network in order to speed up training and to better generalize to new crop
fields.

We achieve that by augmenting the input to the CNN with additional channels
that have previously been used when designing hand-crafted features for
classification~\cite{lottes2016icra} and that provide relevant information for
the classification process. Our approach yields a  pixel-wise semantic
segmentation of the image data and can, given the structure of our network, be
computed at near the frame-rate of a typical camera.

In sum, we make three key claims, which are the following: Our approach is
able to
(i) accurately perform pixel-wise semantic segmentation of crops, weeds, and
soil, properly dealing with heavily overlapping objects and targeting a large
variety of growth stages, without relying on expensive near infra-red
information; 
(ii) act as a robust feature extractor that generalizes well to lighting,
soil, and weather conditions not seen in the training set, requiring little
data to adapt to the new environment; 
(iii) work in real-time on a regular GPU such that operation at near the
frame-rate of a regular camera becomes possible.

%%%%%%%%%%%%%%%%%%%%%%%%%%%%%%%%%%%%%%%%%%%%%%%%%%%%%%%%%%%%%%%%%%%%%%%%%%%%%%%%
%\break
\section{Related Work}
\label{sec:related}
 
% Supervised learning of plants and weeds:
In the context of crop classification, several supervised learning algorithms
have been proposed in the past few years~\cite{haug2014wacv,lottes2016jfr,lottes2016icra,mccool2017ral,milioto2017uavg,mortensen2016cigr,potena2016ias}.
McCool \emph{et al.}~\cite{mccool2017ral} address the crop and weed
segmentation problem by using an ensemble of small CNNs compressed from a more
complex pretrained model and report an accuracy of nearly 94\%. Haug \emph{et
al.}~\cite{haug2014wacv} propose a method to distinguish carrot plants and
weeds in RGB and near infra-red (NIR) images. They obtain an average accuracy
of 94\% on an evaluation dataset of 70 images. Mortensen \emph{et
al.}~\cite{mortensen2016cigr} apply a deep CNN for classifying different types
of crops to estimate their amounts of biomass. They use RGB images of field
plots captured at $3$\,m above ground and report an overall accuracy of 80\%
evaluated on a per-pixel basis. Potena \emph{et al.}~\cite{potena2016ias}
present a multi-step visual system based on RGB+NIR imagery for crop and weed
classification using two different CNN architectures. A shallow network
performs the vegetation detection and then a deeper network further
distinguishes the detected vegetation into crops and weeds. They perform a
pixel-wise classification followed by a voting scheme to obtain predictions
for connected components in the vegetation mask. They report an average
precision of $98$\% in case the visual appearance has not changed between
training and testing phase.

In our previous works~\cite{lottes2016jfr,lottes2016icra}, we presented a
vision-based classification system based on RGB+NIR images for sugar beets and
weeds that relies on NDVI-based pre-segmentation of the vegetation. The
approach combines appearance and geometric properties using a random forest
classifier and obtains classification accuracies of up to 96\% on pixel
level. The works, however, also show that the performance decreases to an
unsuitable level for weed control applications when the appearance of the
plants changes substantially. 
In a second previous work~\cite{milioto2017uavg}, we presented a 2-step
approach that pre-segments vegetation objects by using the NDVI index and
afterwards uses a CNN classifier on the segmented objects to distinguish them
into crops and weeds. We show that we can obtain state-of-the-art object-wise
results in the order of 97\% precision, but only for early grow stages, since
the approach is not able to deal with overlapping plants due to the pre-segmentation step.
 
Lottes \emph{et al.}~\cite{lottes2017iros} address the generalization problem by using
the crop arrangement information, for example from seeding, as prior and exploit
a semi-supervised random forest-based approach that combines this
geometric information with a visual classifier to  quickly adapt to new fields.
Hall \emph{et al.}~\cite{Hall2015wacv} also address the issue of changing
feature distributions. They evaluate different features for leaf classification
by simulating real-world conditions using basic data augmentation techniques.
They compare the obtained performance by
selectively using different handcrafted and CNN features and conclude that CNN
features can support the robustness and generality of a classifier.

In this paper, we also explore a solution to this problem using a CNN-based
feature extractor and classifier for RGB images that uses no geometric prior,
and generalizes well to different soil, weather, and illumination conditions. To
achieve this, we use a combination of end-to-end efficient semantic segmentation
networks and several vegetation indices and preprocessing mappings to the RGB
images.

% Amount of labeling:
A further challenge for these supervised classification approaches is the
necessary amount of labeled data required for retraining to adapt to a new
field. Some approaches have been investigated to address to this problem such
as~\cite{cicco2016arXiv,rainville2012ppa,wendel2016icra}. Wendel and
Underwood~\cite{wendel2016icra} address this issue by proposing a method for
training data generation. They use a multi-spectral line scanner mounted on a field
robot and perform a vegetation segmentation followed by a crop-row detection.
Subsequently, they assign the label crop for the pixels corresponding to the
crop row and the remaining ones as weed. Rainville \emph{et al.}~\cite{rainville2012ppa} 
propose a vision-based method to learn a
probability distributions of morphological features based on a previously
computed crop row. Di~Cicco~\emph{et al.}~\cite{cicco2016arXiv} try to
minimize the labeling effort by constructing synthetic datasets using a
physical model of a sugar beet leaf. Unlike most of these approaches, our
purely visual approach relies solely on RGB images and can be applied to
fields where there is no crop row structure.

\begin{figure}[]
\vspace{2mm} 
\setlength\tabcolsep{1pt}
  \begin{tabular}{ccc}
      {\includegraphics[width=0.49\linewidth]{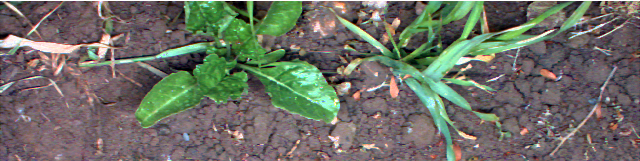}}\vspace{-1mm}
   &
      {\includegraphics[width=0.49\linewidth]{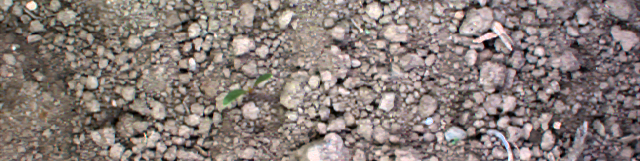}}
   \\ 
      {\includegraphics[width=0.49\linewidth]{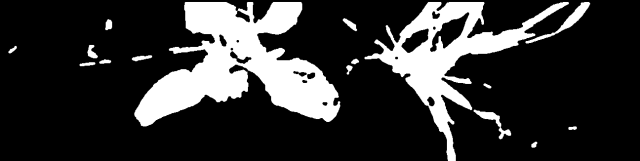}}\vspace{-1mm}
   &
      {\includegraphics[width=0.49\linewidth]{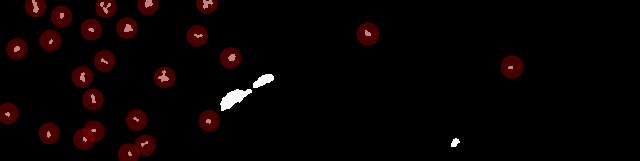}}
   \\
      {\includegraphics[width=0.49\linewidth]{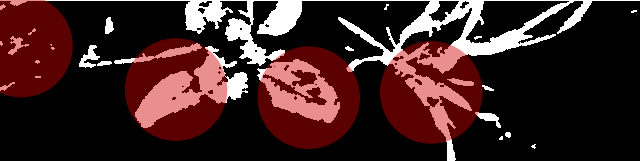}}\vspace{-1mm}
   &
      {\includegraphics[width=0.49\linewidth]{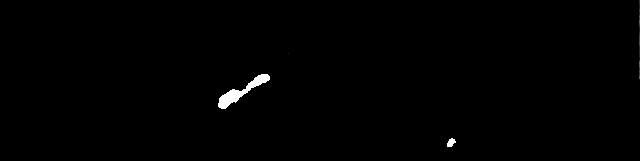}}
  \end{tabular}
\caption{Example image of dataset A (left column) and dataset B (right
column). Top row: RGB image. Middle row: Vegetation mask obtained by applying
a global threshold learned by Otsu's method, which fails for underrepresented vegetation. 
Bottom row: Vegetation detected by adaptive thresholding, which oversegments big plants. 
Failure cases are depicted in red.}
\label{fig:fail}
\end{figure}

%Segmentation vegetation, mask thresholded: 
In addition to the crop weed
classification, several approaches have been developed for vision-based
vegetation detection by using RGB as well as multispectral images, 
e.g.,~\cite{Guo2013cea,hamuda2016cea,lottes2016icra,TorresSanchez2015cea}. 
Often, popular vegetation indices such as the Normalized Difference
Vegetation Index~(NDVI) or the Excess Green Index~(ExG) are used to separate
the vegetation from the background by applying threshold-based approaches.
We see a major problem with these kind of approaches in terms of the
transferability to other fields, i.e. when the underlying distribution of the
indices changes due to different soil types, growth stages, illumination, and
weather conditions. \figref{fig:fail} visually illustrates frequent failure
cases: First, a global threshold, here estimated by Otsu's method, does not properly
separate the vegetation if the plants are in a small growth stage, i.e. when the
vegetation is underrepresented. This problem can be solved with adaptive
thresholding, by tuning the kernel size and a local threshold for small plants.
Second, the adaptive method fails, separating components that should be
connected,  when using the tuned hyperparameters in another dataset where the
growth stage and field condition have changed. Thus, we argue that a learning
approach that eliminates this pre-segmentation and solves segmentation and
classification jointly is needed. As this paper illustrates, such strategy
enables us to learn a semantic segmentation which  generalizes well over several
growth stages as well as field conditions.

%%%%%%%%%%%%%%%%%%%%%%%%%%%%%%%%%%%%%%%%%%%%%%%%%%%%%%%%%%%%%%%%%%%%%%%%%%%%%%%%
 
\section{Approach}
\label{sec:main}

The main goal of our work is to allow agricultural robots to accurately
distinguish weeds from value crops and soil in order to enable intelligent
weed control in real time. We propose a purely visual pixel-wise classifier
based on an end-to-end semantic segmentation CNN that we designed with
accuracy, time and hardware efficiency as well as generalization in mind.

Our classification pipeline can be seen as being separated in two main steps:
First, we compute different vegetation indices and alternate representations
that are commonly used in plant classification and support the CNN with that
information as additional inputs. This turns out to be specifically useful as
the amount of labeled training data for agriculture fields is limited. Second,
we employ a self-designed semantic segmentation network to provide a per-pixel
semantic labeling of the input data. The following two subsections  discuss
these two steps of the pipeline in detail.

\subsection{Input Representations}
\label{sec:representations}
%\FloatBarrier

% \begin{figure}[]  
% \setlength\tabcolsep{2pt}
%   \begin{tabular}{ccc}
%       {\includegraphics[width=0.32\linewidth]{pics/diff_fields/CKA2.png}}
%    &
%       {\includegraphics[width=0.32\linewidth]{pics/diff_fields/Renn.png}}
%    &
%       {\includegraphics[width=0.32\linewidth]{pics/diff_fields/zurich.png}}
%   \end{tabular}
% \caption{Illustration of different weather and soil conditions and growth stages. \todo{Check this legend}}
% \label{fig:data_samples_fields}
% \end{figure}

In general, deep learning approaches make as little assumptions as possible
about the underlying data distribution and let the optimizer decide how to
adjust the parameters of the network by feeding it with enormous amounts of
training data. Therefore, most approaches using convolutional neural networks
make no modifications to the input data and feed the raw RGB images to the
networks.  

% In the particular case of vegetation segmentation, the near infra-red
% information is sometimes used as an extra channel if the sensor can record this
% extra cue.

This means that if we want a deep network to accurately recognize plants and
weeds in different fields, we need to train it with a large amount of data
from a wide variety of fields, lighting and weather conditions, growth stages,
and soil types. Unfortunately, this comes at a high cost and we therefore
propose to make assumptions about the inputs in order to generalize better
given the limited training data.

To alleviate this problem, we provide additional vegetation indexes that have
been successfully used for the plant classification task and that can be
derived from RGB data. In detail, we calculate four different vegetation
indices which are often used for the vegetation segmentation, see
\cite{meyer2008cea,ponti2013}: Excess Green (ExG), Excess Red (ExR), Color
Index of Vegetation Extraction (CIVE), and Normalized Difference Index (NDI).
These indices share the property that they are less sensitive to changes in
the mentioned field conditions and therefore can aid the classification task.
They are computed straightforwardly as:

\small
\begin{align}
\image{ExG} &=  2\,\image{G} - \image{R} - \image{B}\\
\image{ExR} &=  1.4\,\image{R} - \image{G}\\
\image{CIVE} &= 0.881\,\image{G} - 0.441\,\image{R} - 0.385\,\image{B} - 18.78745\\
\image{NDI} &=  \frac{\image{G} - \image{R}}{\image{G} + \image{R}}
\end{align}
\normalsize
\hfill

\begin{figure}[]
\vspace{2mm}  
\setlength\tabcolsep{2pt}
  \begin{tabular}{ccc}
      $\image{RGB}$
   &
      $\image{ExG}$
   &        
      $\image{ExR}$
   \\
      {\includegraphics[width=0.31\linewidth]{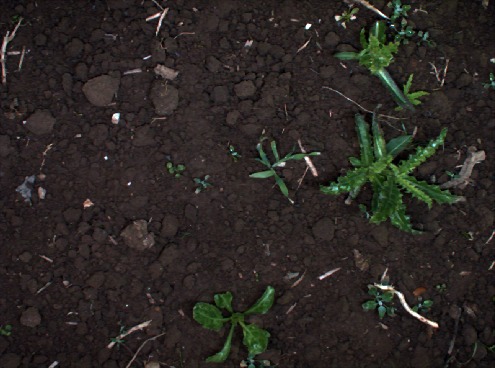}}
   &
      {\includegraphics[width=0.31\linewidth]{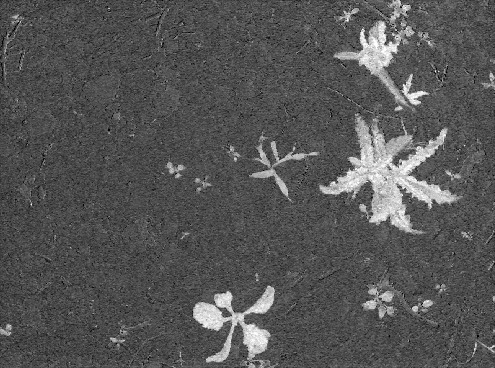}}
   &        
      {\includegraphics[width=0.31\linewidth]{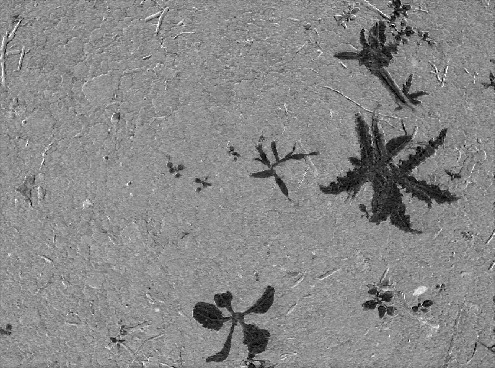}}
   \\
      $\image{HUE}$
   &
      $\image{EDGES}$
   &        
      $\nabla_y\image{ExG}$
   \\
      {\includegraphics[width=0.31\linewidth]{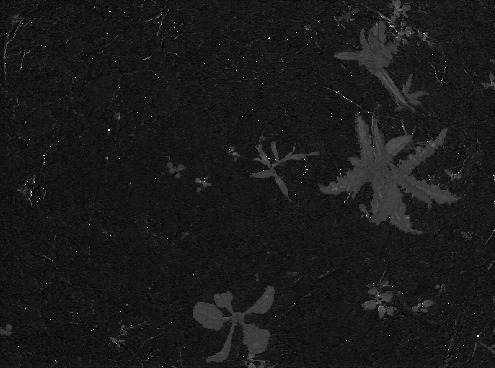}}
   &
      {\includegraphics[width=0.31\linewidth]{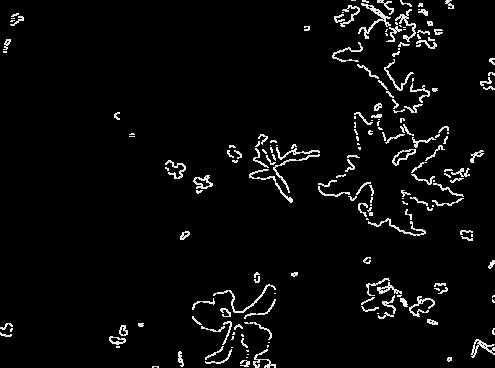}}
   &
      {\includegraphics[width=0.31\linewidth]{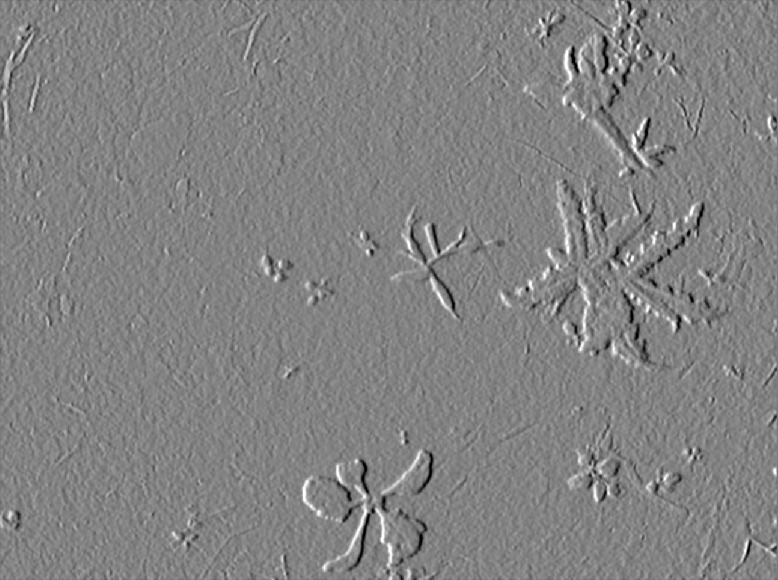}}
  \end{tabular}
\caption{Illustration of some of the used alternate representations.}
\label{fig:feats}
\end{figure}

Along with these four additional cues, we use (i) further representations of
the raw input such as the HSV color space and (ii) operators on the indices
such as the Sobel derivatives, the Laplacian, and the Canny edge detector. All
these representations are concatenated to the channel-wise normalized input
RGB image and build the input volume which is fed into the convolutional
network.  In sum, we use the $14$ channels listed in
\tabref{tab:representations}, and \figref{fig:feats} illustrates how some of
these representations look like. We show in our experiments that deploying
these extra representations to the raw inputs helps not only to learn weight
parameters which lead to a better generalization property of the network, but
also obtain a better performance for separating the vegetation, i.e. crops and
weeds, from soil, and speed up the convergence of the training process.
%This is the case as the amounts of manually labeled training data that is
%available from crop fields is comparably limited.

\begin{table}[]
\caption{Indices and representations used as input by our CNN}
% \vspace{3mm}
\centering
\begin{tabular}{l||l}
  \hline
  \multicolumn{2}{c}{Input channels for our CNN}\\
  \hline
   $\image{1}$ &$\image{R}$\\
   $\image{2}$ &$\image{G}$\\
   $\image{3}$ &$\image{B}$\\
   $\image{4}$ &$\image{ExG}$\\
   $\image{5}$ &$\image{ExR}$\\
   $\image{6}$ &$\image{CIVE}$\\
   $\image{7}$ &$\image{NDI}$\\
   $\image{8}$ &$\image{HUE}$ (from HSV colorspace)\\
   $\image{9}$ &$\image{SAT}$ (from HSV colorspace)\\
   $\image{10}$ &$\image{VAL}$ (from HSV colorspace)\\
   $\image{11}$ &$\nabla_x\image{ExG}$ (Sobel in x direction on $\image{ExG}$)\\
   $\image{12}$ &$\nabla_y\image{ExG}$ (Sobel in y direction on $\image{ExG}$)\\
   $\image{13}$ &$\nabla^2\image{ExG}$ (Laplacian on $\image{ExG}$)\\
   $\image{14}$ &$\image{EDGES}$ (Canny Edge Detector on $\image{ExG}$)\\ \hline
\end{tabular}
\label{tab:representations}
\end{table}

\begin{figure*}
\vspace{2mm}  
\centering
\includegraphics[width=0.95\linewidth]{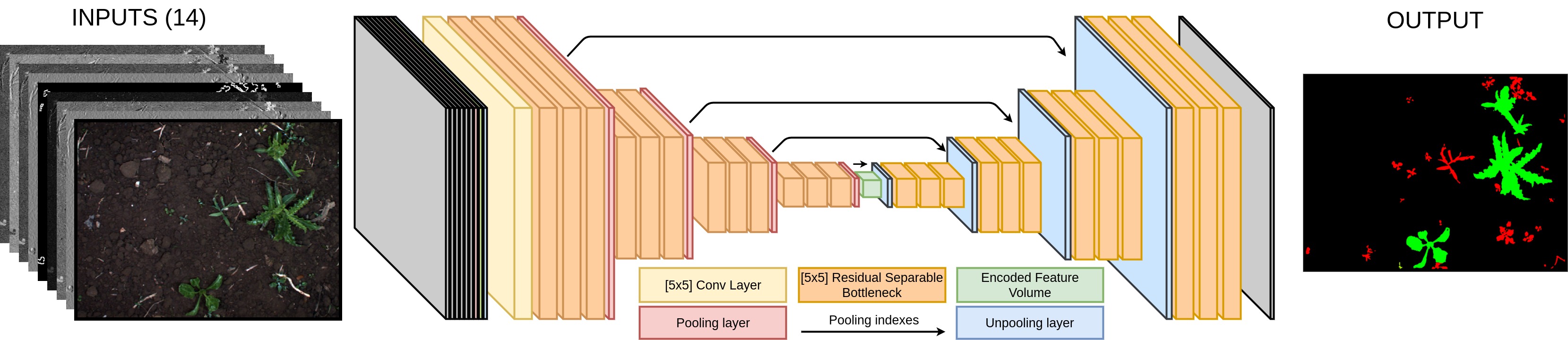}
\caption{Detailed encoder-decoder architecture used for the pixel-wise semantic segmentation. Best viewed in color.}
\label{fig:arch_cnn_3d}
\end{figure*}
 
\subsection{Network Architecture}

Semantic segmentation is a memory and computationally expensive task. 
State-of-the-art algorithms for multi-class pixel-wise classification use CNNs that
have tens of millions of parameters and therefore need massive amounts of
labeled data for training. 
Another problem with these networks is the fact that they often cannot be run 
fast enough for our application. The current state-of-the-art CNN
for this task, Mask R-CNN~\cite{he2017arxiv}, processes around 2 images
per second, whereas we target on a classification rate of at least $10\,\si{Hz}$.
 
Since our specific task of crop vs. weed classification has a much narrower
target space compared to the general architectures designed for $1000$+ classes,
we can design an architecture which meets the targeted speed and
efficiency. We propose an end-to-end encoder-decoder semantic segmentation
network, see \figref{fig:arch_cnn_3d}, that can accurately perform the 
pixel-wise prediction task while running at $20\text{+}\,\si{Hz}$, can be trained 
end-to-end with a moderate size training dataset, and has less than $30{,}000$ parameters.
We design the architecture taking some design cues from
Segnet~\cite{badrinarayanan2015segnet} and Enet~\cite{paszke2016arxiv} into 
account and adapt them to the task at hand. 
Our network is based on the following building blocks:\\

\vspace{-2mm}
\textbf{Input:} We use the representation volume described in
\tabref{tab:representations} as the input to our network. Before it is
passed to the first convolutional layer, we perform a resizing to
$512\times384$ pixels and a channel-wise contrast normalization.

\textbf{Convolutional Layer:} We define our convolutional layers as a
composite function of a convolution followed by a batch normalization and use
the rectified linear unit (ReLU) for the non-linearity. The batch
normalization operation prevents an internal covariate shift and allows for
higher learning rates and better generalization capabilities. The ReLU is
computationally efficient and suitable for training deep networks. All
convolutional layers use zero padding to avoid washing out edge information.

\textbf{Residual Separable Bottleneck:} To achieve a faster processing
time while keeping the receptive field, we propose to use the principal building
block for our network, which is built upon the ideas of (i) residual
connections, (ii) bottlenecks, and (iii) separating the convolutional
operations.
%The key idea is to reduce the number of operations
%and the redundancy.

\begin{figure}[b]  
\centering
\includegraphics[width=0.6\linewidth]{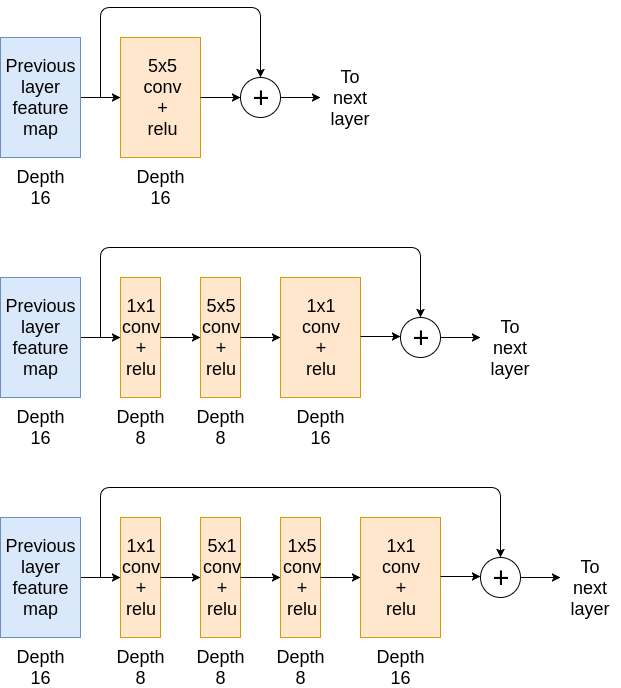}
\caption{Building the residual block to replace a $[5\times5]$ layer.}
\label{fig:arch_resblock}
\end{figure}

\figref{fig:arch_resblock} illustrates the evolution from a conventional
$[5\times5]$ convolutional layer to a residual separable bottleneck. The top
row of \figref{fig:arch_resblock} shows the addition of a residual connection,
which adds the input of the convolution to its result. The addition of this
residual connection helps with the degradation problem that appears when
training very deep networks which makes them obtain a higher training error than
their shallower counterparts~\cite{he2016cvpr}.

The middle row of \figref{fig:arch_resblock} adds $[1\times1]$ convolutions that reduce the depth of the
input volume so that the expensive $[5\times5]$ operation does not need to run
in the whole depth. For this, we use $8\times[1\times1]$ kernels, which halve
the depth of the input volume, and $16\times[1\times1]$ kernels to
expand the result, which needs to match the depth of the input in order to be
added to it. This reduces the amount of calculations per
operation of the kernel from $6{,}400$\,FLOPs to $1{,}856$\,FLOPs.
% ($8\times1\times1\times16+8\times5\times5\times8+16\times1\times1\times8$).

Finally, the bottom row of \figref{fig:arch_resblock} shows the separation of each $[5\times5]$ convolution
into a $[5\times1]$ convolution followed by a $[1\times5]$ convolution. This
further reduces the operations of running the module in a $[5\times5]$
window from $1{,}856$\,FLOPs to $896$\,FLOPs.
% ($8\times1\times1\times16+2\times(8\times5\times1\times8)+16\times1\times1\times8$). 

These design choices also decrease the number of parameters for the
equivalent layer to the $[5\times5]$ convolution with 16 kernels from $6{,}400$
to $896$ parameters.

\textbf{Unpooling with Shared Indexes:} The unpooling operations in the
decoder are performed sharing the pooling indexes of the symmetrical pooling
operation in the encoder part of the network. This allows the network to
maintain the information about the spatial positions of the maximum
activations on the encoder part without the need for transposed convolutions,
which are comparably expensive to execute. Therefore, after each unpooling
operation, we obtain a sparse feature map, which the subsequent convolutional
layers learn how to densify. All pooling layers are $[2\times2]$ with stride $2$.
 
\textbf{Output:} The last layer is a linear operation followed by a softmax
activation predicting a pseudo-probability vector of length $3$ per pixel,
where each element represents the probability of the pixel belonging to the
class background, weed, or crop.\\
\vspace{-2mm}

We use these building blocks to create the network depicted in
\figref{fig:arch_cnn_3d}, which consists of an encoder-decoder architecture. The
encoder part has $13\times[5\times5]$ convolutional layers containing $16$
kernels each and $4$ pooling layers that reduce the input representation into a
small code feature volume. This is followed by a decoder with $12$ convolutional
layers containing $16$ kernels each and $4$ unpooling layers that upsample this
code into a semantic map of the same size of the input. The design of the
architecture is inspired by Segnet's design simplicity, but in order to make it
smaller, easier to train, and more efficient, we replace $24$ of the $25$
convolutional layers by our proposed residual separable convolutional
bottlenecks. The $5\times5$ receptive fields of the convolutional layers, along
with the pooling layers add up to an equivalent receptive field in the input
image plane of $200\times200$ pixels. This is sufficient for our application, since it
is a bigger window than the biggest plant we expect in our data. The proposed
number of layers, kernels per layer, and reduction factor for each bottleneck
module was chosen by training several networks with different configurations,
and reducing its size until we reached the smallest configuration that did not
result in a considerable performance decrease.

%%%%%%%%%%%%%%%%%%%%%%%%%%%%%%%%%%%%%%%%%%%%%%%%%%%%%%%%%%%%%%%%%%%%%%%%%%%%%%%%
\section{Experimental Evaluation}
\label{sec:exp}

\begin{table}[]
\vspace{2mm}  
\centering
\caption{Dataset Information}
\begin{tabular}{|r||c|c|c|}
    \hline
   & \textbf{Bonn} & \textbf{Zurich} & \textbf{Stuttgart} \\ \hline 
   \#~images & 10,036 &  2,577 & 2,584 \\\hline 
   \#~crops & 27,652 & 3,983 & 10,045 \\\hline
   crop pixels & 1.7\% & 0.4\% & 1.5\% \\\hline
   \#~weeds & 65,132 & 14,820 & 7,026 \\\hline 
   weed pixels & 0.7\% & 0.1\% & 0.7\% \\\hline 
  \end{tabular}
\label{tab:datasets}
\end{table}

The experiments are designed to show the accuracy and efficiency of our method
and to support the claims made in the introduction that our classifier is able
to perform accurate pixel-wise classification of value crops and weeds,
generalize well, and run in real-time.

We implemented the whole approach presented in this paper relying on the Google
TensorFlow library, and OpenCV to compute alternate representations of the
inputs. We tested our results using a Bosch Deepfield Robotics BoniRob UGV.

\subsection{Training and Testing Data}

\begin{figure}[]
\vspace{2mm}  
\setlength\tabcolsep{4pt}
  \begin{tabular}{ccc}
      \small{Bonn} & \small{Zurich} & \small{Stuttgart} \\
      {\includegraphics[width=0.30\linewidth]{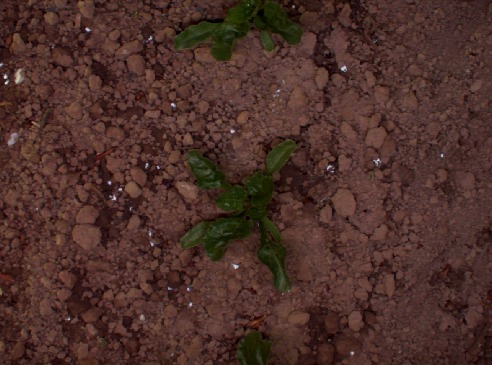}}
   &
      {\includegraphics[width=0.30\linewidth]{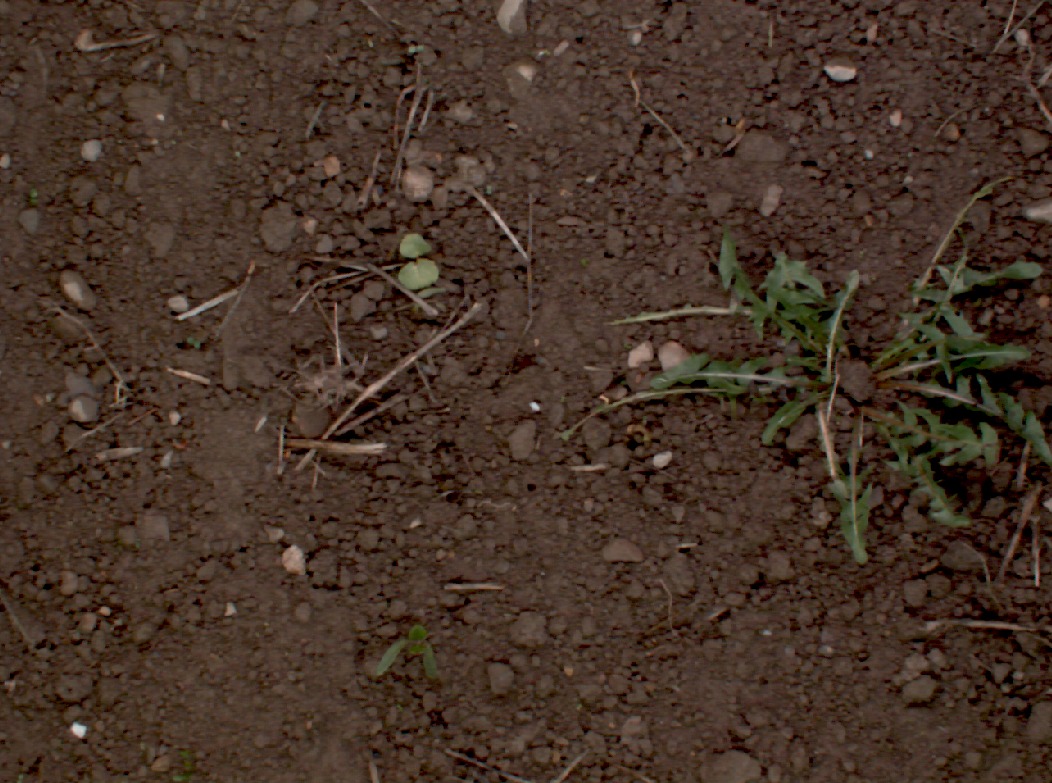}}
   &
      {\includegraphics[width=0.30\linewidth]{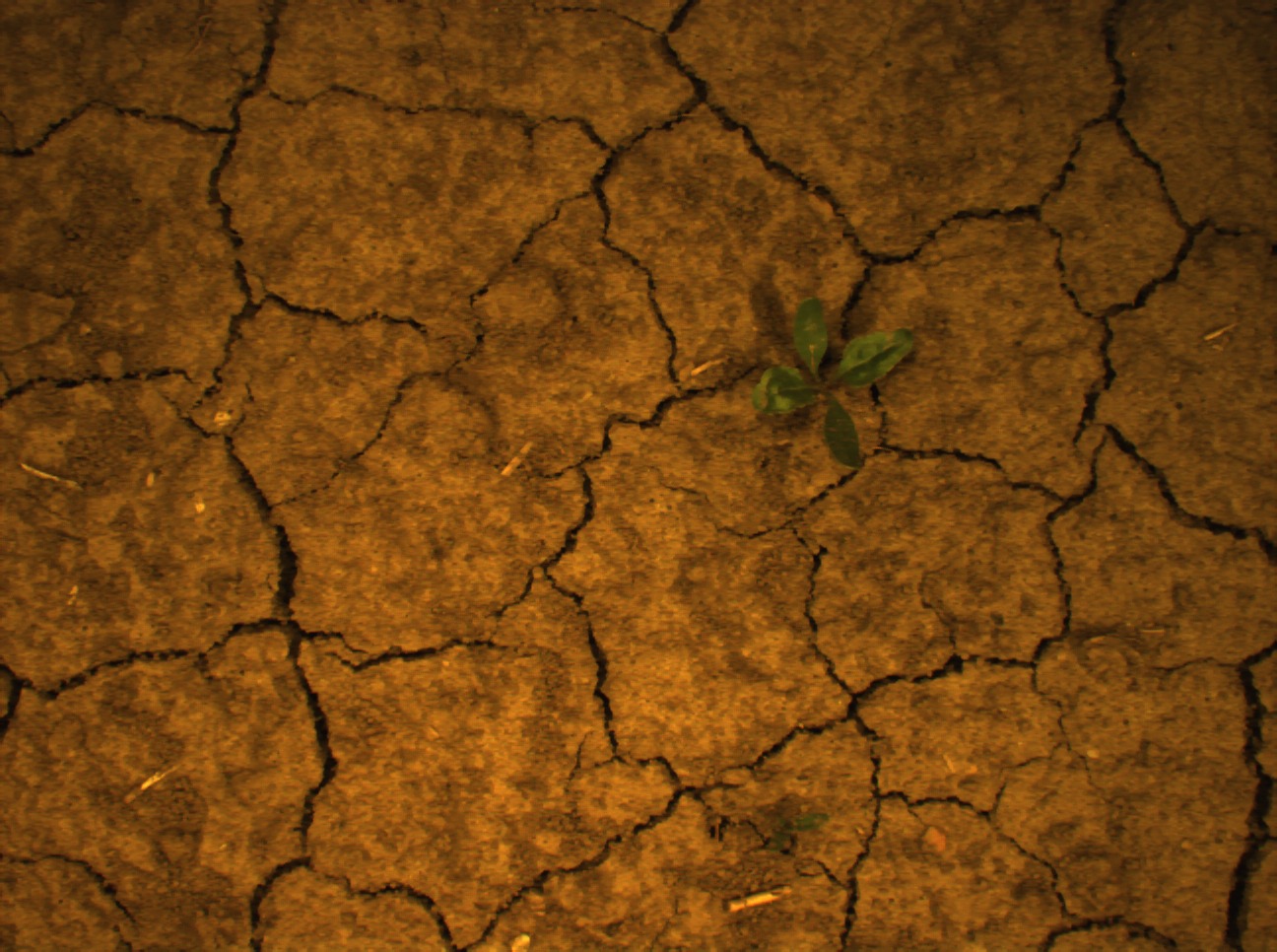}}
  \end{tabular}
\caption{Example images from each dataset. Every dataset contains different
crops growth stages, weed types, illumination conditions and soil types. Best viewed in color}
\label{fig:data_samples_fields}
\end{figure}

To evaluate the performance of the network, we use three different datasets
captured in Bonn, Germany; Stuttgart, Germany; and Zurich, Switzerland, see
\tabref{tab:datasets}. 
Part of the data from Bonn is publicly available~\cite{chebrolu2017ijrr}.
All datasets contain plants and weeds in
all growth stages, with different soil, weather, and illumination conditions.
\figref{fig:data_samples_fields} illustrates the variance of the mentioned
conditions for each dataset. The visual data was recorded with the
4-channel RGB+NIR camera JAI AD-130 GE mounted in nadir view. For our approach,
we use solely the RGB channels because we aim to perform well with an off-the-shelf
RGB camera, but the additional NIR information allows us to compare the performance
with an approach that exploits the additional NIR information.

To show the generalization capabilities of the pipeline, we train our network
using only images from Bonn, captured over the course of one month, and
containing images of several growth stages. We separate the dataset in
70\%-15\%-15\% for training, validation, and testing, and we report our results
only in the latter 15\% held out testing portion, as well as the whole of the
two other datasets from Zurich and Stuttgart, some even recorded in different
years.

We train the network using the 70\% part of the Bonn dataset extracted for that
purpose and perturb the input data by performing random rotations, scalings,
shears and stretches. We use Stochastic Gradient Descent with a batch size of 15
for each step, which is the maximum we can fit in a single GPU.
Furthermore, we use a weighted cross-entropy loss, to handle the imbalanced
number of pixels of each class, due to the soil dominance, and the Adam
optimizer for the calculation of the gradient steps. Training for 200 epochs
takes roughly 48 hours on an NVIDIA GTX1080Ti.

To show the effect in performance obtained by the extra channels,
we train three networks using different types of inputs: one based solely on RGB
images, another one based on RGB and extra representations, and finally the
reference baseline network using RGB+NIR image data.

%%%%%%%%%%%%%%%%%%%%%%%%
\subsection{Performance of the Semantic Segmentation}

% \emph{Start EVERY experiment with a similar start as the following
%   sentence, explaining in the first sentence why you present the
%   experiment and which claim it aims at supporting.}

% %% First experiment - most impressive, important or the most important
% %% claim supporting experments comes first.
% The first experiment is designed to show the performance of our
% approach and to support the claim that it is well-suited for ....

% \begin{figure}[t]
%   \centering
%  \includegraphics[width=0.99\columnwidth]{./pics/motivation.png}
%   \caption{A caption that makes you understand the image easily.}
%   \label{fig:parameval}
% \end{figure}

% %% Second experiment - could be a comparison to a baseline methods,
% %% quality analysis or similar
% The second experiment is to support the claim that our approach is ...

%%%%%%%%%%%%%%%%%%%%%%%%

This experiment is designed to support our first claim, which states that our 
approach can accurately perform pixel-wise semantic segmentation of crops, weeds,
and soil, properly dealing with heavy plant overlap in all growth stages.

\begin{table*}[]
\vspace{2mm}
\caption{Pixel-wise Test Sets Performance. Trained in 70\% Bonn, reporting 15\% held-out Bonn, 100\% Stuttgart, and 100\% Zurich.}
\centering
\begin{tabular}{r|c|cccccccccc}
    \hline
      \multirow{2}{*}{Dataset} & \multirow{2}{*}{Network} & \multirow{2}{*}{mIoU[\%]} & \multicolumn{3}{c}{IoU[\%]} & \multicolumn{3}{c}{Precision[\%]} & \multicolumn{3}{c}{Recall[\%]}\\ 
      &  & & Soil & Weeds & Crops & Soil & Weeds & Crops & Soil & Weeds & Crops \\
      \hline
      \multirow{3}{*}{Bonn} & $\image{RGB}$ & 59.98 & 99.08 & 20.64 & 60.22 & 99.92 & 28.97 & 66.49 & 99.15 & 41.79 & 82.45 \\
      & $\image{RGB}$+$\image{NIR}$ & 76.92 & 99.29 & 49.42 & 82.06 & 99.88 & 52.90 & 84.19 & 99.33 & \textbf{88.24} & 97.01 \\
      & $\image{1}...\image{14}$ (ours) & \textbf{80.8} & \textbf{99.48} & \textbf{59.17} & \textbf{83.72} & \textbf{99.95} & \textbf{65.92} & \textbf{85.71} & \textbf{99.53} & 85.25 & \textbf{97.29} \\
      \hline
      \hline
      \multirow{3}{*}{Zurich} & $\image{RGB}$ & 38.25 & 96.84 & 14.26 & 3.62 & 96.95 & 14.96 & 3.78 & 96.88 & 35.35 & 45.86 \\
      & $\image{RGB}$+$\image{NIR}$ & 41.23 & 98.44 & 16.83 & 8.43 &  99.68 & 19.03 & 9.07 & 98.46 & \textbf{51.27} & 54.46 \\
      & $\image{1}...\image{14}$ (ours) & \textbf{48.36} & \textbf{99.27} & \textbf{23.40} & \textbf{22.39} & \textbf{99.90} & \textbf{31.43} & \textbf{23.05} & \textbf{99.36} & 47.79 & \textbf{88.74} \\
      \hline
      \multirow{3}{*}{Stuttgart} & $\image{RGB}$ & 48.09 & 99.18 & 21.40 & 23.69 & 99.84 & 21.90 & 52.43 & 99.34 & 42.95 & 28.95 \\
      & $\image{RGB}$+$\image{NIR}$ & 55.82 & 98.54 & 23.13 & 45.80 & 99.85 & 25.28 & 68.76 & 98.69 & \textbf{49.10} & 57.84\\
      & $\image{1}...\image{14}$ (ours) & \textbf{61.12} & \textbf{99.32} & \textbf{26.36} & \textbf{57.65} & \textbf{99.86} & \textbf{37.58} & \textbf{68.77} & \textbf{99.45} & 46.90 & \textbf{78.09} \\
      \hline
\end{tabular}
\label{tab:pixel_test}
\end{table*}

In \tabref{tab:pixel_test}, we show the pixel-wise performance of the
classifier tested in the 15\% held out test set from Bonn, as well as the
whole datasets from Zurich and Stuttgart. These results show that the network
trained using the RGB images in conjunction with the extra computed
representations of the inputs outperforms the network using solely RGB images
significantly in all categories, performing comparably with the network that
uses the extra visual cue from the NIR information, which comes with a high
additional cost as specific sensors have to be employed.

Even though the pixel-wise performance of the classifier is important, in order
to perform automated weeding it is important to have an object-wise metric for
the classifier's performance. We show this metric in \tabref{tab:obj_test},
where we analyze all objects with area bigger than 50 pixels. This number is
calculated by dividing our desired minimum object detection size of
$1~\si{cm^2}$ by the spatial resolution of $2~\si{mm^2}/\text{px}$ in our
$512\times384$ resized images. We can see that in terms of object-wise
performance the network using all representations outperforms its RGB
counterpart. Most importantly, in the case of generalization to the datasets of
Zurich and Stuttgart, this difference becomes critical, since the RGB network
yields a performance so low that it renders the classifier unusable for most
tasks.

Note that the network using RGB and extra representations is around 30\% faster
to converge to 95\% of the final accuracy than its RGB counterpart, and roughly
15\% faster than the network using the NIR channel.

\small
\tabcolsep=0.11cm
\begin{table}[]
%\vspace{2mm}
\caption{Object-wise Test Performance. Trained in 70\% Bonn, reporting 15\% held-out Bonn, 100\% Stuttgart, and 100\% Zurich.}
\centering
\begin{tabular}{r|c|ccccc}
    \hline
      \multirow{2}{*}{Dataset} & \multirow{2}{*}{Network} & \multirow{2}{*}{mAcc[\%]} & \multicolumn{2}{c}{Precision[\%]} & \multicolumn{2}{c}{Recall[\%]}\\ 
      & & & Weeds & Crops & Weeds & Crops \\
      \hline
      \multirow{3}{*}{Bonn} & $\image{RGB}$ & 86.34 & 83.63 & 81.14 & 91.99 & 80.42 \\
      & $\image{RGB}$+$\image{NIR}$ & 93.72 & 90.51 & \textbf{95.09} & \textbf{94.79} & 89.46 \\
      & $\image{1}...\image{14}$ (ours) & \textbf{94.74} & \textbf{98.16} & 91.97 & 93.35 & \textbf{95.17} \\
      \hline
      \hline
      \multirow{3}{*}{Zurich} & $\image{RGB}$ & 45.51  & 59.75  & 19.71  & 42.52  & 23.66 \\
      & $\image{RGB}$+$\image{NIR}$ & 68.03 & 67.41 & 46.78 & \textbf{65.31} & 49.32 \\
      & $\image{1}...\image{14}$ (ours) & \textbf{72.08}  & \textbf{67.91}  & \textbf{72.55}  & 63.33  & \textbf{64.94} \\
      \hline
      \multirow{3}{*}{Stuttgart} & $\image{RGB}$ & 46.05 & 42.32 & 42.03 & 46.1 & 25.01 \\
      & $\image{RGB}$+$\image{NIR}$ & 73.99 & 74.30 & \textbf{70.23} & \textbf{71.35} & 53.88 \\
      & $\image{1}...\image{14}$ (ours) & \textbf{76.54} & \textbf{87.87} & 65.25 & 64.66 & \textbf{85.15} \\
      \hline
\end{tabular}
\label{tab:obj_test}
\end{table}
\normalsize

\subsection{Labeling Cost for Adaptation to New Fields}

This experiment is designed to support our second claim, which states that our approach
can act as a robust feature extractor for images in conditions not seen in the
training set, requiring little data to adapt to the new environment.

One way to analyze the generalization performance of the approach is to analyze the amount of data
that needs to be labeled in a new field for the classifier to achieve
state-of-the-art performance. For this, we separate the Zurich and Stuttgart datasets
in halves, and we keep 50\% of it for testing. From each of the remaining
50\%, we extract sets of 10, 20, 50 and 100 images, and we retrain the last layer
of the network trained in Bonn, using the convolutional layers in the encoder and
the decoder as a feature extractor.  
We further separate this small sub-samples in 80\%-20\% for training and 
validation and we train until convergence, using early stopping, which means that
we stop training when the validation error starts to increase.
This is to provide an automated approach to the retraining, so that it can be 
done without supervision of an expert.

\begin{figure}[h]
\vspace{2mm}   
  \begin{center}
   \includegraphics[width=0.85\linewidth]{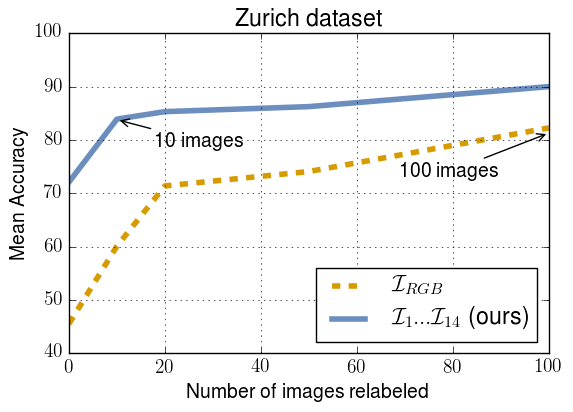}\\
   \vspace{7mm}
   \includegraphics[width=0.85\linewidth]{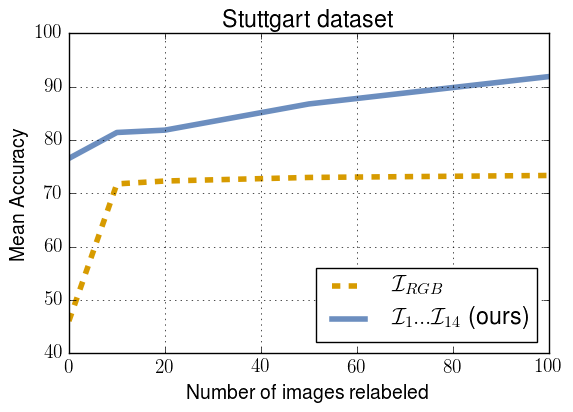}
  \end{center}
\caption{Object-wise mean accuracy vs. relabeling effort. Zero images means no retraining.}
\label{fig:relabel}
\end{figure}

We show the results of the retraining on the Zurich dataset in
\tabref{tab:retrain_zurich} and \figref{fig:relabel}, where we can see
that the performance of the RGB network when relabeling 100 images is
roughly the same as the one using all input  representations when the latter is
using only 10 images, thus significantly reducing the relabeling effort. We also
show that we can get values of precision and recall in the order of 90\% when
using 100 images for the relabeling in the case of our network, which exploits
additional channels. In \tabref{tab:retrain_stuttgart} and
\figref{fig:relabel}, we show the same for the Stuttgart dataset, but
in this case the RGB network fails to reach an acceptable performance, while the
accuracy of our approach grows linearly with the number of images used.

\tabcolsep=0.11cm
\begin{table}[t]
{\small
\caption{Object-wise test performance retraining in N images of Zurich dataset.}
%\vspace{2mm}
\centering
\begin{tabular}{r|c|c|ccccc}
    \hline
      \multirow{2}{*}{Inputs} & \multirow{2}{*}{Nr. Images} & \multirow{2}{*}{mAcc[\%]} & \multicolumn{2}{c}{Precision[\%]} & \multicolumn{2}{c}{Recall[\%]}\\ 
      & & & Weeds & Crops & Weeds & Crops \\
      \hline
      \multirow{4}{*}{$\image{RGB}$} & 10 & 60.08 & 58.75 & 62.22 & 73.03 & 57.99 \\
      & 20 & 71.38 & 61.38 & 81.42 & 76.72 & 64.58 \\
      & 50 & 74.08 & 63.00 & 85.42 & 74.14 & 68.34 \\
      & 100 & 82.26 & 65.50 & 85.59 & 74.97 & 69.91 \\
      \hline
      \multirow{4}{1cm}{\centering $\image{1}...\image{14}$ (ours)} & 10 & 83.90 & 69.23 & 80.73 & 71.71 & 76.18 \\
      & 20 & 85.31 & 75.85 & 79.12 & 67.67 & 84.01 \\
      & 50 & 86.25 & 76.50 & 85.24 & 71.55 & 84.33 \\
      & 100 & \textbf{89.55} & \textbf{85.89} & \textbf{89.52} & \textbf{89.69} & \textbf{86.76} \\
      \hline
\end{tabular}
\label{tab:retrain_zurich}
%\end{table}
%\normalsize
\vspace{3mm}
%\small
%\tabcolsep=0.11cm
%\begin{table}[h]
\caption{Object-wise Test performance retraining in N images of Stuttgart dataset.}
%\vspace{2mm}
\centering
\begin{tabular}{r|c|c|ccccc}
    \hline
      \multirow{2}{*}{Inputs} & \multirow{2}{*}{Nr. Images} & \multirow{2}{*}{mAcc[\%]} & \multicolumn{2}{c}{Precision[\%]} & \multicolumn{2}{c}{Recall[\%]}\\ 
      & & & Weeds & Crops & Weeds & Crops \\
      \hline
      \multirow{4}{*}{$\image{RGB}$} & 10 & 71.76 & 93.88 & 52.59 & 56.57 & 81.10 \\
      & 20 & 72.30 & 94.40 & 57.72 & 51.43 & 86.35 \\
      & 50 & 72.97 & 94.33 & 59.70 & 54.11 & 87.69 \\
      & 100 & 73.34 & 95.20 & 63.26 & 56.36 & 87.66 \\
      \hline
      \multirow{4}{1cm}{\centering $\image{1}...\image{14}$ (ours)} & 10 & 81.40 & 89.48 & 64.42 & 78.74 & 79.69 \\
      & 20 & 81.84 & 87.83 & 68.18 & 81.45 & 74.61 \\
      & 50 & 86.75 & 94.68 & 71.34 & 83.22 & 90.23 \\
      & 100 & \textbf{91.88} & \textbf{95.45} & \textbf{86.58} & \textbf{89.08} & \textbf{91.43} \\
      \hline
\end{tabular}
\label{tab:retrain_stuttgart}
}
\end{table}
%\normalsize

% \begin{figure}[h]  
% \setlength\tabcolsep{2pt}
%       {\includegraphics[width=0.7\linewidth]{pics/graphs/zurich_retrain.png}}
% \caption{Mean Accuracy vs Relabeling Effort, Zurich Dataset.\\ Zero images means
% no retraining}
% \label{fig:relabel_zurich}
% \end{figure}

% \begin{figure}[h]  
% \setlength\tabcolsep{2pt}
%       {\includegraphics[width=0.7\linewidth]{pics/graphs/stuttgart_retrain.png}}
% \caption{Mean Accuracy vs Relabeling Effort, Stuttgart Dataset.\\ Zero images means
% no retraining}\label{fig:relabel_stuttgart}
% \end{figure}

%%\FloatBarrier

\subsection{Runtime}
This experiment is designed to support our third claim, which states that our approach
can be run in real-time, and therefore it can be used for online operation in
the field, running in hardware that can be fitted in mobile robots.

We show in \tabref{tab:runtime} that even though the architecture using all
extra representations of the RGB inputs has a speed penalty, due to the efficiency
of the network we can run the full classifier at more than 20 frames per second on
the hardware in our UGV, which consists of an Intel i7 CPU and an NVIDIA 
GTX1080Ti GPU. Furthermore, we tested our approach in the Jetson TX2 platform, 
which has a very small footprint and only takes $15\,\si{W}$ of peak power, making it
suitable for operation on a flying vehicle, and here we still obtain a frame rate of almost
$5\,\si{Hz}$.

\small
\tabcolsep=0.11cm
\begin{table}[h]
\caption{Runtime in different devices.}
%\vspace{2mm}
\centering
\footnotesize
\begin{tabular}{r|c|c|cc|c|c}
  \hline
  Inputs & FLOPS & Hardware & Preproc. & Network & Total & FPS\\
  \hline
  \multirow{2}{*}{RGB} & \multirow{2}{*}{$1.8\text{G}$} & i7+GTX1080Ti & - & $31\si{ms}$ & $31\si{ms}$ & $32.2$\\
  & & Tegra TX2 SoC & - & $190\si{ms}$ & $190\si{ms}$ & $5.2$\\
  \hline
  \multirow{2}{*}{All} & \multirow{2}{*}{$2\text{G}$} & i7+GTX1080Ti & $6\si{ms}$ & $38\si{ms}$ & $44\si{ms}$ & $22.7$\\
  & & Tegra TX2 SoC & $6\si{ms}$ & $204\si{ms}$ & $210\si{ms}$ & $4.7$\\
  \hline
\end{tabular}
\label{tab:runtime}
\end{table}
\normalsize

\section{Conclusion}
\label{sec:conclusion}

In this paper, we presented an approach to pixel-wise semantic segmentation of 
crop fields  identifying  crops, weeds, and background in real-time  solely from RGB data.
We proposed a deep encoder-decoder CNN for semantic segmentation that is fed with a 14-channel
image storing vegetation indexes and other information that in the past has been used to solve
crop-weed classification tasks.
By feeding this additional, task-relevant background knowledge to the network, we can 
speed up training and  improve the generalization capabilities on
new crop fields, especially if the amount of training data is limited.
We implemented and thoroughly evaluated our system on a real agricultural robot operating 
using data from three different cities in Germany and Switzerland.  
Our results suggest that our system generalizes well, can operate at around 20\,Hz, 
and is suitable for online operation in the fields.
 
%%%%%%%%%%%%%%%%%%%%%%%%%%%%%%%%%%%%%%%%%%%%%%%%%%%%%%%%%%%%%%%%%%%%%%%%%%%%%%%%
% Only if applicable
\section*{Acknowledgments}
We thank the teams from the Campus Klein Altendorf in Bonn and ETH Zurich Lindau-Eschikon
 for supporting us and maintaining the fields. We furthermore thank 
BOSCH Corporate Research and DeepField Robotics for their support during 
joint experiments. Further thanks to N.~Chebrolu for collecting and 
sharing  datasets~\cite{chebrolu2017ijrr}.

\bibliographystyle{plain}
\bibliography{glorified}

\end{document}